\pdfoutput=1

\documentclass[11pt]{article}

\usepackage[]{acl}

\usepackage{times}
\usepackage{latexsym}

\usepackage[T1]{fontenc}

\usepackage[utf8]{inputenc}

\usepackage{microtype}

\usepackage{inconsolata}

\usepackage{hyperref}

\usepackage{graphicx}
\usepackage{array}
\usepackage{arydshln}
\usepackage{booktabs}
\usepackage{amsmath}
\usepackage{multirow}
\usepackage{multicol}
\usepackage{algorithm} 
\usepackage{algorithmic} 
\title{Enhancing Model Privacy in Federated Learning with \\ Random Masking and Quantization}

\author{Zhibo Xu\thanks{Equal contribution.}, \
{\bf Jianhao Zhu\footnotemark[1]}, \
{\bf Jingwen Xu}, \
{\bf Changze Lv}, \
{\bf Zhenghua Wang}, \\
{\bf Zisu Huang}, \
{\bf Xiaohua Wang}, \
{\bf Muling Wu}, \
{\bf Qi Qian}, \\
{\bf Xiaoqing Zheng\thanks{Corresponding Author.}}, \
{\bf Xuanjing Huang} \\
School of Computer Science, Fudan University, Shanghai, China \\
\texttt{$\{$zbxu23,zhujh22$\}$@m.fudan.edu.cn}\\
\texttt{$\{$zhengxq,xjhuang$\}$@fudan.edu.cn}
}

\begin{document}
\maketitle
\begin{abstract}
The primary goal of traditional federated learning is to protect data privacy by enabling distributed edge devices to collaboratively train a shared global model while keeping raw data decentralized at local clients.
The rise of large language models (LLMs) has introduced new challenges in distributed systems, as their substantial computational requirements and the need for specialized expertise raise critical concerns about protecting intellectual property (IP). This highlights the need for a federated learning approach that can safeguard both sensitive data and proprietary models.
To tackle this challenge, we propose FedQSN, a federated learning approach that leverages random masking to obscure a subnetwork of model parameters and applies quantization to the remaining parameters. Consequently, the server transmits only a privacy-preserving proxy of the global model to clients during each communication round, thus enhancing the model’s confidentiality.
Experimental results across various models and tasks demonstrate that our approach not only maintains strong model performance in federated learning settings but also achieves enhanced protection of model parameters compared to baseline methods. Code and resources are available at \href{https://github.com/zb2313/FedQSN}{https://github.com/zb2313/FedQSN}.
\end{abstract}

\section{Introduction}

Federated learning (FL) is a distributed machine learning paradigm that preserves data privacy by enabling a central server to train a model without direct access to clients’ raw data~\citep{nguyen2022federated,long2020federated,imteaj2021survey,lim2020federated}.  A canonical example is FedAvg~\citep{mcmahan2017communication}, in which the server broadcasts global parameters, clients perform local updates, and the server aggregates the returned updates.

Meanwhile, large language models (LLMs) have become valuable intellectual property: they acquire rich semantic knowledge through large-scale pretraining and underpin commercial APIs such as OpenAI’s GPT series~\citep{radford2019language} and Google’s PaLM~\citep{chowdhery2023palm}.  Because FedAvg shares full model parameters each round, it risks exposing proprietary LLM weights and undermining IP protection—thereby motivating methods that guard model confidentiality in FL.  

In federated learning scenarios where an internet company collaborates with multiple hospitals to train a comprehensive medical model, both parties face distinct privacy concerns. The internet company, having invested substantial resources in training a large language model, seeks to protect its intellectual property by preventing the full exposure of its model parameters and capabilities. Meanwhile, the hospitals aim to safeguard the privacy of their sensitive local medical data. This dual-privacy setting has driven recent advances in federated learning research, with growing attention to protecting server-side model privacy in addition to client-side data privacy. 

To address the risk of intellectual property leakage, several approaches have been proposed that preserve model privacy—such as constructing proxy models—while maintaining the ability to protect user data. FedSP~\cite{dong2023tunable} is the first work to address large language model (LLM) privacy in federated learning. It introduces a proxy model mechanism, allowing clients to download a surrogate model instead of accessing the server’s global model parameters. However, FedSP requires the server to possess a labeled dataset that is independently and identically distributed (IID) with client data—a condition rarely met in practice and potentially at odds with the privacy goals of federated learning.
FedLPP~\cite{zhu2024promoting} eliminates the need for such auxiliary datasets by sharing only quantized LoRA (Low-Rank Adaptation) adapters with clients, thus enhancing model privacy. Nonetheless, the backbone model must still be distributed to clients before training begins. If this model contains proprietary information, it may expose the server’s intellectual property. Additionally, FedLPP offers limited protection for the server-side model.

Previous studies have shown that training only a subset of a model's parameters can yield performance comparable to training the full model~\cite{houlsby2019parameter,xu2024random}. While quantization is commonly used to reduce model size and communication overhead for deployment on resource-constrained platforms\cite{xiao2023smoothquant,yao2023comprehensive}, it essentially works by lowering the precision of model parameters. This reduction in precision makes it more difficult for external parties to recover exact parameter values, thereby helping preserve the model owner's competitive advantage and protect the intellectual property of the model\cite{zhu2024promoting,colombo2025quantization}. Building on these concepts, we introduce the FedQSN approach, which combines random masking and quantization to enhance model privacy protection while maintaining model performance with minimal degradation.

\begin{figure*}[t]
  \centering
  \includegraphics[width=0.95\linewidth]{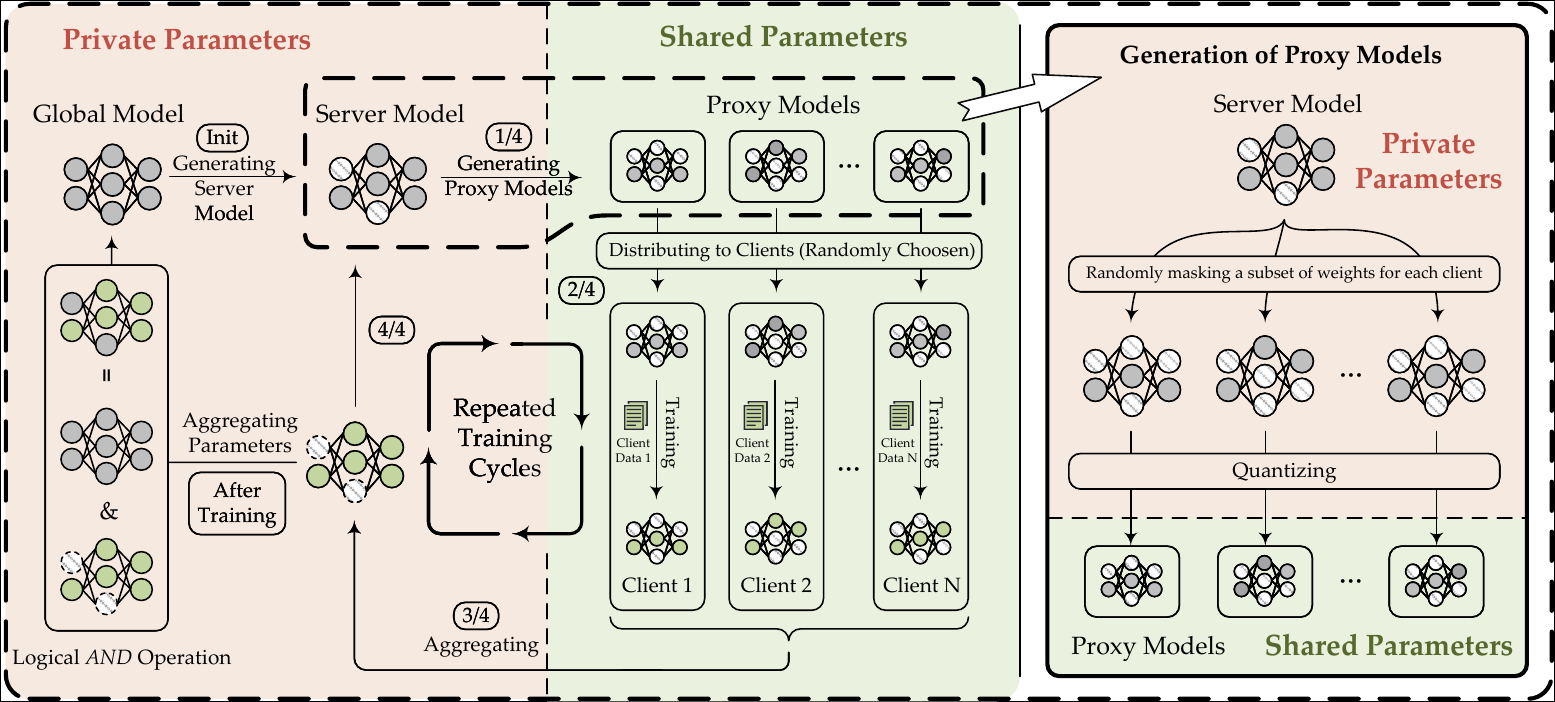} \hfill
  
  \caption{The proposed federated learning approach protects both server-side model and client-side data privacy. (1) The global model is first masked with a server-side mask to create the "server model." In each round, this server model is perturbed with client-specific masks and quantized to form the "proxy model." (2) The proxy model is sent to clients for local training. (3) After training, clients return parameter updates, which are aggregated by the server. (4) The updated server model serves as the starting point for the next round. After all training rounds, the proxy model and the original global model are combined via a logical \textit{AND} operation to produce the final global model, enhancing both privacy and performance.}
  
  \label{fig:main}
\end{figure*}

In the proposed FedQSN framework, the server first applies a global random mask to hide a subset of model parameters from all clients, preventing any client coalition from reconstructing the complete model. Then, a client-specific random mask is applied to each selected client to further increase the masking ratio for that round. Due to randomized client selection and masking in each communication round, every client trains a unique subnetwork on its private data, while ensuring that all model parameters are updated sufficiently over time. Prior to transmission, the masked model is quantized to lower-precision representations, which not only obscures exact parameter values but also reduces communication overhead. This process produces a privacy-preserving proxy model for client-side training. After all training rounds, the final model is reconstructed by applying a logical AND operation between the trained model and the original untrained model retained by the server, effectively restoring the masked parameters. While federated learning inherently safeguards data privacy, FedQSN offers additional protection for model confidentiality. A schematic overview is provided in Figure~\ref{fig:main}. Experimental results show that FedQSN significantly enhances model privacy with minimal performance degradation compared to existing baselines.

The major contributions of this paper are summarized as follows:

\begin{itemize}

\item We propose FedQSN, a novel federated learning algorithm that simultaneously preserves the privacy of the server-side model and the client-side data. Clients interact only with a proxy model, which undergoes server-side random masking and quantization, instead of accessing the complete global model.

\item Extensive experiments conducted on diverse datasets and model architectures demonstrate that our method significantly improves model privacy compared to baseline approaches, while maintaining competitive performance without significant degradation.

\item We conduct comprehensive ablation studies to assess the individual and combined effects of each component in our proposed framework.
\end{itemize}
\section{Related Work}
\subsection{Traditional Federated Learning}
Federated Learning (FL)~\cite{konecny2016federated, mcmahan2017communication, yang2019federated, kairouz2021advances} enables multiple clients to train a shared model without sharing raw data. The application of FL in the field of Natural Language Processing (NLP) addresses the issues of data leakage and privacy infringement caused by traditional centralized model training methods. Previous related work has primarily focused on addressing heterogeneity~\cite{li2019convergence, zhao2018federated, jeong2018communication}, communication~\cite{shahid2021communication, konevcny2016federated, luping2019cmfl}, robustness~\cite{bonawitz2017practical} and data privacy protection~\cite{bogdanov2008sharemind, geyer2017differentially, cai2020under} issues.

\subsection{Federated Learning for LLMs}
However, in the context of large language models (LLMs), traditional federated learning (FL) approaches face scalability challenges, primarily due to communication and training inefficiencies caused by the transmission of full model parameters. To address this, FedRDMA~\citep{zhang2024fedrdma} leverages RDMA-based communication by splitting model updates into blocks for more efficient transfer. FedPETuning~\citep{zhang2023fedpetuning} explores parameter-efficient fine-tuning (PEFT) methods to improve training efficiency in FL, while~\citep{wang2024personalized} proposes a gradient-free prompt tuning approach for better few-shot performance. In terms of privacy, FedML-HE~\citep{jin2023fedml} introduces selective homomorphic encryption to protect sensitive parameters with reduced overhead.

\subsection{Model privacy protection in FL}

To prevent the intellectual property of NLP models from being leaked to clients participating in Federated Learning, some studies have begun to explore model protection privacy in Federated Learning.

FedSP~\cite{dong2023tunable} introduces a proxy model and a tunable soft prompt as intermediaries between the server and clients, avoiding direct sharing of the global model. To mitigate the performance gap between the proxy and global models, FedSP requires the server to possess an auxiliary dataset that is independent and identically distributed (IID) with the client data. However, this assumption contradicts the core principle of federated learning—protecting client data privacy—as the auxiliary dataset may reveal sensitive features to the server. Moreover, constructing such a dataset is often impractical in real-world scenarios.
FedLPP~\cite{zhu2024promoting} improves model protection by decomposing the global model into a backbone and a LoRA adapter, sharing only the backbone and a quantized proxy adapter with clients. While this design prevents direct exposure of the global adapter, it only protects a portion of the model parameters and thus does not achieve full model-level privacy.

In contrast, another line of work focuses on data-level privacy. Methods such as EW-Tune\cite{behnia2022ew}, Whispered Tuning\cite{singh2024whispered}, Split-N-Denoise\cite{mai2023split} (SnD), and user-level DP-SGD\cite{charles2024fine}, focus on protecting data privacy through techniques like differential privacy, PII redaction, client-side noise addition, and sampling strategies. These methods are orthogonal to ours, which focuses on model-level privacy through random masking and quantization, rather than data-level protection. Thus, our approach can complement data-centric techniques, as DP mechanisms cannot substitute parameter-level defenses.

\section{Method}
In the following sections, we will introduce the improved federated learning approach we proposed. We will sequentially present the random masking and model quantization methods introduced to enhance model privacy protection, as well as the overall training process.

\subsection{Preliminary}
\label{sec:preliminary}
We first revisit the fundamental objective of federated learning (FL), which aims to minimize the global loss function:
\begin{equation}
\min_{\mathbf{W}} \mathcal{L}(\mathbf{W}) = \sum_{k=1}^{N} \frac{|\mathcal{D}_k|}{\sum_{k=1}^N|\mathcal{D}_k|} \mathcal{L}_k(\mathbf{W})
\label{equation:loss}
\end{equation}
where $\mathbf{W}$ represents the model parameters, $|\mathcal{D}_k|$ denotes the size of the private dataset for the $k$-th client, $\mathcal{L}_k(\cdot)$ is the local loss function for dataset $\mathcal{D}_k$, and $N$ is the total number of clients. The primary challenge lies in training a unified model across distributed clients while ensuring the protection of data privacy.
\subsection{Random Model Masking}
\label{sec:random-mask}
To protect the privacy of the server-side model by limiting client access to its full parameters, we introduce a random mask applied to the global model before it is transmitted from the server to the clients during federated learning training. Specifically, we inject a column-wise random mask directly into the model weights of every attention and MLP layer before distribution. Let $W$ be a weight matrix and let $M$ be a random mask whose entries are sampled independently as in Equation \ref{eqa:mask1}:
\begin{equation}
    M_{:,j}=\{\begin{array}{cc}0&\text{with probability }p\\\frac{1}{1-p}&\text{with probability }1-p\end{array}
    \label{eqa:mask1}
\end{equation}
where $M$ has the same dimensions as $W$, the masked weights are obtained by a Hadamard product, refer to Equation \ref{eqa:mask2}:
\begin{equation}
    \widetilde{W}=W\odot M
    \label{eqa:mask2}
\end{equation}
with probability $1-p$ the column in matrix $W$ is kept but amplified by $\frac{1}{1-p}$, so that $E[\widetilde{W}]=W$ and the expected magnitude of activations is preserved. This adjustment follows the same principle as Dropout ~\cite{srivastava2014dropout}, ensuring that the overall distribution of the outputs remains stable, preventing significant shifts due to the masking.

During training, the server employs a hierarchical masking mechanism to safeguard the model before distributing its parameters to the clients. This process involves two sequential masking steps: first, a server-side mask, referred to as the server mask, is applied to the global model. Then, a client-specific mask, which we refer to as the client mask, is applied when the model is distributed to each client. A detailed description of this dual-masking strategy can be found in Section \ref{sec:training_procedure}.

\subsection{Model Quantization}

\label{sec:model-quantization}
Model quantization, a widely used compression technique, reduces model size by lowering parameter precision, often at the expense of performance. By quantizing the models distributed from the server to clients, we hinder accurate reconstruction of server parameters, thereby enhancing model privacy. In this work, we employ conventional blockwise quantization with $\omega$-bit precision, as established in prior studies \cite{dettmers2024qlora,zhu2024promoting}.

The quantization process begins by selecting a target bit-width $\omega$, which determines the quantization levels. The floating-point values in the weight matrix $W$ are then partitioned into blocks of size $s$, preserving spatial continuity in the flattened representation. For each block $X_i$, we compute the maximum absolute value, denoted $absmax(X_i)$. The quantized parameters of the $i$-th block, $\hat{X_i}$, are then obtained by scaling and rounding to the nearest integer, as expressed in Equation \ref{eqn:xl}:
\begin{equation}
    \hat{X_i}=round(\frac{2^{\omega-1}-1}{absmax(X_i)}X_i) \label{eqn:xl}
\end{equation}

\begin{table*}[htb]
  \centering
  \begin{tabular}{llccccc}
    \toprule[1.3pt]
    \multicolumn{2}{c}{\textbf{Method}}&\textbf{BLEU}&\textbf{NIST}&\textbf{METEOR}&\textbf{ROUGE-L}&\textbf{CIDEr}\\
    \midrule[1.3pt]
    {\textbf{FedAVG}}      & global &35.04 & 6.58 &32.64& 52.01& 1.80    \\
    \midrule
    \multirow{1}{*}{\textbf{FedSP } }& global & 26.42& 3.65& 25.88 &44.42 &1.21  \\
                
    \midrule
    \multirow{3}{*}{\textbf{FedLPP } }     
                & global &  34.60& 6.06&31.43&	51.32 &	1.70    \\
                \multirow{1}{*}{} & proxy & 32.46 & 5.12 & 29.72 & 50.31 & 1.52  \\
     \multirow{1}{*}{} & gap & 2.14 & 0.94 & 1.71 & 1.01 & 0.18  \\
    \midrule
    \multirow{3}{*}{\textbf{FedQSN } } 
                & global & 33.91&5.52 &30.70	&49.27  & 1.61
    \\
        \multirow{1}{*}{}& proxy & 25.15 &  3.37 & 23.76&42.56 & 1.01
 \\
       \multirow{1}{*}{}& gap &\textbf{ 8.76} & \textbf{2.15} & \textbf{6.94} & \textbf{6.71 }& \textbf{0.60}
 \\
    \bottomrule[1.3pt]
  \end{tabular}
  \caption{
   Comparative evaluation of FedQSN against three baseline methods, with metric values averaged across four benchmark datasets (E2E, DART, DIALOGSUM, ViGGO).
  }
  \label{tab:compare_with_baselines}
\end{table*}

\begin{figure*}[t] 
  \centering
  \includegraphics[width=1\linewidth]{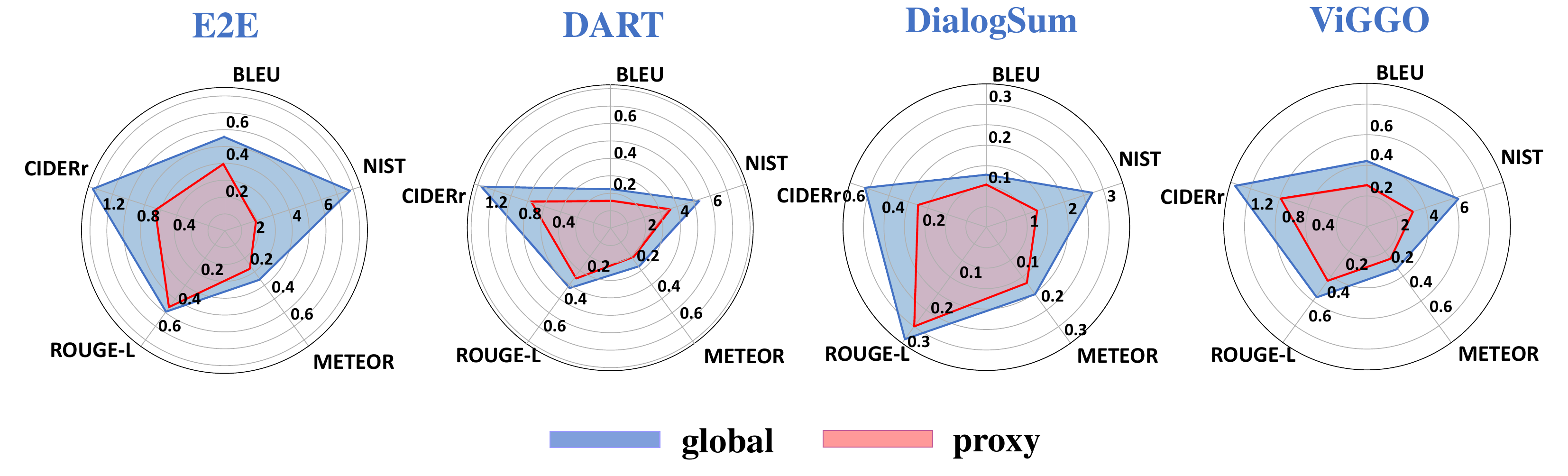} \hfill
  
  \caption {Performance of global and proxy models across multiple metrics for FedQSN on various datasets.}
  \label{fig:radar}
\end{figure*}
\begin{table*}[htb]
  \centering
  \small
  \setlength{\tabcolsep}{15pt}
  \begin{tabular}{ccccccc}
    \toprule[1.3pt]
    \multicolumn{2}{c}{\textbf{Method}}&\textbf{BLEU}&\textbf{NIST}&\textbf{METEOR}&\textbf{ROUGE-L}&\textbf{CIDEr}\\
    \midrule[1.3pt]
    \multirow{1}{*}{\textbf{FedAVG } } 
                &global &56.40& 7.78&41.07&62.64& 1.64   \\
    \midrule
    \multirow{2}{*}{\textbf{FedQSN }$\omega = 1$ } 
                & global &39.90& 4.46& 22.35& 47.67& 0.59    \\
       
       \multirow{2}{*}{} & proxy&8.29& 0.02&7.83& 27.48& 0.02   \\
    \midrule
    \multirow{2}{*}{\textbf{FedQSN }$\omega = 2 $ } 
                & global & 52.14& 6.46& 31.38& 56.72& 1.31   \\
       
       \multirow{2}{*}{} & proxy &24.74& 0.35& 19.34& 45.69& 0.53
     \\
    \midrule
    \multirow{2}{*}{\textbf{FedQSN }$\omega = 3 $ } 
                & global & 54.18& 7.32&33.41& 58.41&1.46 \\
       
       \multirow{2}{*}{} & proxy &41.56&2.12&25.09& 53.60&0.85     \\
    \midrule
    \multirow{2}{*}{\textbf{FedQSN }$\omega = 4 $ } 
                & global& 54.98&7.49&33.70&58.54&1.50     \\
       
       \multirow{2}{*}{} & proxy & 48.18 &3.60&27.14&55.67&1.04  \\
    \bottomrule[1.3pt]
  \end{tabular}
  \caption{\label{table:compare_different_quantization}
    Performance of our method with varying bit width ($\omega$) settings. To maintain consistent experimental conditions, the server mask ratio is fixed at $p_1 = 0.1$ and the client mask ratio at $p_2 = 0.1$. 
    Evaluations are performed on the E2E dataset using the GPT-2 Medium model to assess the impact of different bit widths on performance.
  }
\end{table*}

\begin{table*}[htb]
  \centering
  \small
  \begin{tabular}{ccccccc}
    \toprule[1.3pt]
    \multicolumn{2}{c}{\textbf{Method}}&\textbf{BLEU}&\textbf{NIST}&\textbf{METEOR}&\textbf{ROUGE-L}&\textbf{CIDEr}\\
    \midrule[1.3pt]
    \multirow{1}{*}{\textbf{FedAVG } } 
                &global &56.40& 7.78&41.07&62.64& 1.64   \\
    \midrule
    \multirow{2}{*}{\textbf{FedQSN }$p_1 = 0.1,p_2 = 0$  } 
    
                & global & 54.57&7.65&35.26& 59.07& 1.62  \\
                
    \multirow{2}{*}{}& proxy &47.50 & 3.42 & 27.05 & 55.48 & 1.04 
    \\
    \midrule
    \multirow{2}{*}{\textbf{FedQSN }$p_1=0,p_2 = 0.1$ }     
                & global &  54.67&7.71 & 35.04 & 59.09 & 1.59    \\
                
     \multirow{2}{*}{}  & proxy &47.72 & 3.47&27.15&56.20&1.03   \\
    \midrule
    \multirow{2}{*}{\textbf{FedQSN }$p_1 = 0.1,p_2 = 0.1$ } 
                & global &55.46 & 7.74 & 34.91&58.48&1.54    \\
                
      \multirow{2}{*}{} & proxy&46.41&3.04&26.52&55.36& 1.01  \\
    \midrule
                
                
    \multirow{2}{*}{\textbf{FedQSN }$p_1 = 0.1,p_2 = 0.2$ } 
                & global &56.02& 7.82& 36.12& 59.50& 1.67    \\
       
       \multirow{2}{*}{} & proxy &48.89&3.63&27.13&56.11&1.07    \\
    \midrule
    \multirow{2}{*}{\textbf{FedQSN }$p_1 = 0.2,p_2 = 0.1 $ } 
                & global & 53.01&6.89 & 31.75 & 56.81 & 1.37     \\
       
       \multirow{2}{*}{} & proxy &27.23&2.40&20.66&48.53&0.60
     \\
    \midrule
    \multirow{2}{*}{\textbf{FedQSN }$p_1= 0.2,p_2 = 0.2 $ } 
                & global &  52.78& 7.24 & 32.54& 57.23&1.45 \\
       
       \multirow{2}{*}{} & porxy & 40.64& 3.07& 26.65& 53.55&0.97   \\

    \bottomrule[1.3pt]
  \end{tabular}
  \caption{\label{table:compare_different_mask}
    Performance of the proposed method under varying random mask settings, with server and client mask ratios $p_1$ and $p_2$ detailed in Section \ref{sec:random-mask}. Experiments use a quantization bit width $\omega = 2$ on the E2E dataset with the GPT-2 Medium model.
  }
\end{table*}
\begin{table*}[htb]
  \centering
  \small
  \begin{tabular}{llccccc}
    \toprule[1.3pt]
    \multicolumn{2}{c}{\textbf{Model}}&\textbf{BLEU}&\textbf{NIST}&\textbf{METEOR}&\textbf{ROUGE-L}&\textbf{CIDEr}\\
    \midrule[1.3pt]

    \multirow{2}{*}{\textbf{GPT2-XL } } 
                & global  & 55.70& 7.66& 42.33& 64.44& 1.68\\
    \multirow{2}{*}{}& proxy &   38.26  & 1.89& 23.53&51.42&0.71   \\
    \midrule
    \multirow{2}{*}{\textbf{Llama3.2-1B } }     

         & global  &  56.19    &  7.85   & 35.37 & 58.72  &  1.63  \\

    \multirow{2}{*}{}& proxy &  40.52 & 1.93 & 25.76 & 54.84 & 0.85      \\
    \midrule
    \multirow{2}{*}{\textbf{Llama3.2-3B } } 
                & global &    56.25     &    7.87  &   36.29    &  59.32  &  1.69   \\
                
       \multirow{2}{*}{} & proxy &    39.46    &  1.86    &   25.49    & 55.27   & 0.90 \\
    \midrule
    \multirow{2}{*}{\textbf{Llama3.1-8B } } 
                & global &    56.45	& 7.80 &	35.49 &	60.42 &	1.70   \\
                
       \multirow{2}{*}{} & proxy &       38.59&	1.92&	24.25&	47.14&	1.21 \\
       
    \bottomrule[1.3pt]
  \end{tabular}
  \caption{\label{compare_different_model_scales}
    Performance of the FedQSN method on the E2E dataset across language models of varying parameter sizes and types, showcasing the versatility of our approach. To ensure consistent experimental conditions, the server mask ratio is fixed at $p_1 = 0.1$, the client mask ratio at $p_2 = 0.1$, and the quantization bit width at $\omega = 2$.
  }
\end{table*}
\begin{table*}[htb]
  \centering
  \small
  \begin{tabular}{llccccc}
    \toprule[1.3pt]
    \multicolumn{2}{c}{\textbf{Method}}&\textbf{BLEU}&\textbf{NIST}&\textbf{METEOR}&\textbf{ROUGE-L}&\textbf{CIDEr}\\
    \midrule[1.3pt]
    \multirow{1}{*}{\textbf{FedAVG(cross silo) } } 
                & global &56.88&7.87& 39.55& 61.93& 1.77\\

    \midrule
    \multirow{2}{*}{\textbf{FedQSN(cross silo) } } 
                & global &56.20&7.71& 39.24& 61.43& 1.65\\

     \multirow{2}{*}{} & proxy &  40.02&2.34& 25.05& 53.43& 0.84   \\
    \midrule
    \multirow{1}{*}{\textbf{FedAVG(large scale)} } 
                & global & 55.61&7.73& 40.76& 62.30& 1.57 \\

    \midrule
    \multirow{2}{*}{\textbf{FedQSN(large scale)} } 
                & global  &  52.14 & 6.45&  31.38 &  56.72& 1.31\\
    \multirow{2}{*}{}& proxy & 24.74 & 0.35 & 19.34 & 45.69 & 0.53 \\

    \bottomrule[1.3pt]
  \end{tabular}
  \caption{\label{table:different_silo}
    This table shows experimental results for federated learning on the E2E dataset using GPT-2 Medium. The "cross silo" scenario uses all 5 clients, while the "large scale" scenario randomly selects 5 from 25 clients. Server and client mask ratios are fixed at $p_1 = p_2 = 0.1$, with quantization bit width $\omega = 2$.
  }
\end{table*}

\subsection{Training Procedure}
\label{sec:training_procedure}
In the proposed FedQSN approach, we first generate a server mask based on the methodology described in Section \ref{sec:random-mask} and apply it to the original global model with a masking proportion $p_1$. This ensures that the masked parameters are inaccessible to clients, preventing full model reconstruction and reducing the risk of collusion among adversarial clients. Next, $C$ clients are randomly selected from the $N$ participants. For each selected client, a client-specific mask with a proportion $p_2$ is generated and applied to the already server-masked model, which increases the overall masking and ensures that each client trains a distinct subnetwork on its private data in each round. Over successive training rounds this guarantees that all parameters receive updates. The doubly masked model is then subjected to block-wise $\omega$-bit quantization (as detailed in Section \ref{sec:model-quantization}) before being distributed to the clients. Quantization obscures the exact parameter values and reduces communication overhead. The resulting privacy-preserving proxy model is then distributed to clients for local training.

During local training, clients iteratively update the received proxy model based on their local datasets. The local optimization process follows the update rule:
\begin{equation}
\label{eqa:local-training}
\mathbf{W}_{t+1} = \mathbf{W}_t + \sum_{k=1}^N \frac{|\mathcal{D}_k|}{\sum_{k=1}^N|\mathcal{D}_k|} \Delta\mathbf{W}_{t}^{k}
\end{equation}
where $\mathbf{W}_t$ denotes the global model parameters at iteration t, $\Delta\mathbf{W}_{t}^{k}$ represents the parameter updates from the k-th client, and the weighting term $\Delta\mathbf{W}_{t}^{k}$ ensures proportional contribution based on client dataset sizes. This federated averaging mechanism aligns with the overarching objective of minimizing the global loss in Equation \ref{equation:loss}.
Upon completing local training, clients transmit $\Delta\mathbf{W}_{t}^{k}$ to the server. The server aggregates the received updates using a weighted summation, as defined in Equation \ref{eqa:local-training}, and applies a secure aggregation protocol \cite{bonawitz2017practical} to ensure privacy during the aggregation process. This results in the updated global model $\mathbf{W}_{t+1}$, which will be used in the subsequent training round. The complete workflow encompassing masking, quantization, and secure aggregation is formalized in Algorithm \ref{alg:combined}. At the conclusion of all training rounds, a logical AND operation is performed between the final aggregated model and the original untrained model retained by the server, effectively restoring the masked parameters and reconstructing the fully trained global model.

\section{Experiments}
In this section, we present an extensive set of experiments to evaluate the effectiveness of our algorithm in preserving model privacy, while maintaining strong performance in federated learning (FL) across a range of tasks. We provide comparative performance results against baseline models.

Our experiments primarily utilize the GPT-2 Medium model in conjunction with four widely recognized datasets: E2E~\cite{novikova2017e2e}, DART~\cite{nan2020dart}, ViGGO~\cite{juraska2019viggo}, and DialogSum~\cite{chen2021dialogsum}. Detailed descriptions of these datasets can be found in Appendix \ref{sec:appendix-datasets}.

For evaluation, we employ several standard metrics, including BLEU~\cite{papinesi2002bleu}, NIST~\cite{Belz_Reiter_2006}, METEOR~\cite{banerjee2005meteor}, ROUGE-L~\cite{chin2004rouge}, and CIDEr~\cite{vedantam2015cider}.


\subsection{Baselines}
We use the following methods as baselines to compare with our proposed FedQSN. To ensure a fair and systematic comparison between different federated learning methods, we conducted the evaluations under consistent settings. The final performance metrics were reported on the test set. For detailed implementation specifics, please refer to Appendix \ref{sec:appendix-experiments}.
\begin{itemize}

    \item \textbf{FedAVG}: Aggregates client-trained global models to preserve data privacy but lacks privacy mechanisms, exposing model leakage risks.
    
    \item \textbf{FedSP}: Pioneers joint model-data privacy via proxy distillation but violates FL principles by requiring client data access during distillation.

    \item \textbf{FedLPP}: Enhances FedAVG with quantized LoRA transmission to reduce gradient leakage. However, partial server-model exposure persists, limiting full privacy. FedLPP balances privacy and efficiency, serving as the key reference for FL privacy advancements.

\end{itemize}

\subsection{Performance Comparison}

Table \ref{tab:compare_with_baselines} presents a comparative analysis between our proposed method, FedQSN, and existing baselines. To evaluate server-side model privacy protection, we follow the evaluation protocol proposed in FedLPP \cite{zhu2024promoting}, which uses the performance gap between the server-held global model and the client-received proxy model as a measure of privacy preservation. Specifically, for both FedLPP and FedQSN, we compare the best-performing global model and the best-performing proxy model across all training rounds. This choice reflects a realistic and worst-case scenario, as both the server and client are expected to retain their most effective models for deployment or further use. If the global model consistently outperforms the proxy model, it indicates effective protection of the server-side model. Moreover, a larger performance gap suggests that the proxy model remains less competitive, providing stronger evidence of privacy preservation within the federated learning framework. The reported results represent the average performance across all four datasets for each method.

As shown in Table \ref{tab:compare_with_baselines}, FedQSN achieves more effective model privacy protection compared to FedLPP. While FedSP also aims to protect model privacy, its global model performance is significantly lower than that of both FedLPP and FedQSN. Across all metrics, the global model trained with FedQSN consistently outperforms the proxy model, indicating that the server retains a performance advantage over individual clients and thus protects the model owner’s proprietary knowledge. Furthermore, the performance of the global model trained with FedQSN is comparable to that of FedAVG, suggesting that our method preserves model utility while enhancing privacy protection. Table \ref{tab:compare_with_baselines} provides a quantitative comparison of the baseline methods based on average results across all datasets.

For detailed performance results of FedQSN on individual datasets, refer to Figure \ref{fig:radar}. As illustrated in Figure \ref{fig:radar}, the global model consistently outperforms the proxy model across all metrics and datasets. Notably, the performance curves for the global model encompass those of the proxy model in all dimensions, empirically validating the strong privacy-preserving capabilities of our method through its robust performance.

We further evaluated the privacy protection of FedQSN by measuring the cosine similarity between the global and proxy model parameters, in comparison with FedLPP. As shown in Table~\ref{tab:para_similarity}, FedQSN exhibits notably lower similarity, indicating superior model privacy protection. Additional details are provided in Appendix~\ref{sec:para-similarity}.

Our method is compatible with secure aggregation protocols \cite{bonawitz2017practical} and does not require a labeled dataset on the server side, thereby avoiding potential risks to client data privacy—unlike FedSP. Consequently, it achieves a level of client data protection comparable to that of FedAVG.

\begin{table}[htb]
\small
    \centering
    \begin{tabular}{cc}
    \toprule[1.3pt]
      \textbf{Method}   & \textbf{Parameter Similarity} \\
      \midrule[1.3pt]
       \textbf{FedLPP}  &0.995 \\
       \midrule
       \textbf{FedQSN}  & 0.805 \\
    \bottomrule[1.3pt]
    \end{tabular}
    \caption{Comparison of parameter similarity between the global model and its proxy under FedQSN and FedLPP.}
    \label{tab:para_similarity}
\end{table}

\section{Ablation Analysis}
In this section, we present a series of ablation studies to evaluate the impact of varying random masking and quantization levels within the FedQSN approach. We also conduct comparison experiments across models of different scales and foundational architectures, as well as experiments in diverse federated learning scenarios.

\subsection{Impact of quantization levels}
FedQSN involves a trade-off between model privacy and performance, governed by the quantization bit width $\omega$. Lower bit widths (smaller $\omega$) enhance privacy by increasing the deviation between the proxy and global models, but also introduce greater gradient bias, reducing performance.

We evaluated bit widths of 1 to 4, with results shown in Table~\ref{table:compare_different_quantization}. Bit widths of 3 and 4 strike a favorable balance, achieving strong privacy protection with minimal performance loss. In contrast, overly aggressive quantization (e.g., $\omega=1$ or 2) leads to significant performance degradation due to insufficient proxy model information, despite stronger privacy guarantees. All experiments were conducted under identical training settings for fair comparison.

\subsection{Impact of random masking levels}
Table~\ref{table:compare_different_mask} shows that both server-side and client-side random masking provide effective privacy protection for the global model in federated learning. Under consistent quantization levels, random masking yields negligible degradation in global model performance while substantially reducing the accuracy of proxy models reconstructed by adversaries. However, excessive masking can significantly impair aggregated model performance, undermining the primary goal of maintaining high accuracy in federated learning.

Additionally, Table~\ref{tab:w/o_mask} in Appendix~\ref{sec:appendix-experiments} compares the individual effects of quantization and random masking. While quantization alone offers stronger privacy protection, it leads to greater degradation in global model performance. In contrast, increasing the masking ratio at a fixed quantization level enhances privacy with only a minor impact on performance.

\subsection{Impact of model scales and various federated learning scenarios}

We extended FedQSN from GPT-2 Medium to larger architectures, including GPT-2 XL~\cite{radford2019language}, Llama3.2-1B, Llama3.2-3B, and Llama3.1-8B~\cite{grattafiori2024llama}. Results on the E2E dataset~\cite{novikova2017e2e}, shown in Table~\ref{compare_different_model_scales}, confirm that FedQSN consistently protects model privacy while maintaining strong performance across model scales.

To assess its robustness in different federated learning (FL) settings, we evaluated FedQSN under two representative scenarios: cross-silo FL, where all clients participate in each round~\cite{kairouz2021advances}, and large-scale cross-device FL, where only a subset of clients with smaller datasets participate due to communication or availability constraints~\cite{lai2022fedscale}.

Experiments using the E2E dataset under both scenarios (see Table~\ref{table:different_silo}) demonstrate that FedQSN maintains effectiveness despite limited client participation, as evidenced by the observed performance gap between global and proxy models. These results highlight FedQSN’s adaptability to diverse and realistic FL environments.

\section{Conclusion}

In this study, we introduce FedQSN, a federated learning approach that enhances the privacy of model parameters. The server applies random masking and quantization to the model parameters before transmitting them to clients, thereby enhancing parameter-level confidentiality. Unlike baseline methods, FedQSN limits clients’ access to the full model and mitigates the risk of reverse-engineering. Experimental results show that FedQSN achieves comparable performance to baseline methods while significantly enhancing server-side model protection. This establishes a clear distinction between the server-side global model and the privacy-preserving proxy models delivered to clients.

\section*{Limitations}

We have demonstrated the effectiveness of FedQSN in protecting model privacy within the federated learning framework across various datasets and models. However, several areas warrant further investigation. First, applying FedQSN to larger-scale models could help assess its scalability and effectiveness in more complex scenarios. Second, evaluating the method on more diverse and complex datasets could offer deeper insights into its robustness in real-world settings. Third, integrating differential privacy—a standard technique for protecting model intellectual property from training data—with our approach could further strengthen model owners’ control in federated learning, representing a promising direction for future research. Addressing these limitations could improve the broader applicability of FedQSN and facilitate its adoption in diverse federated learning scenarios.
\bibliography{custom}
\clearpage
\appendix
\section{Method}
\label{sec:appendix-method}
\begin{algorithm}[htb]
    \caption{\textbf{FedQSN.} R is the number of training rounds, C is the number of chosen clients, E is the number of local epochs, and $\eta$ is the learning rate. $p_1$ and $p_2$ represent the proportions of the server mask and client mask, respectively, while $\omega$ denotes the level of quantization.}
    \label{alg:combined}
    \begin{algorithmic}[1]
        \STATE \textbf{ServerExcute:}
        \STATE initial global model parameters $W_0$
        \STATE $W_0 \leftarrow$ SetModelMask $(W_0,p_1)$
        \STATE $S \leftarrow$ (random choose C clients) 
        \FOR{each round $t = 1,2,...R-1$}
            \FOR{each client $c \in S$ \textbf{in parallel}}
                \STATE $W^c_{t} \leftarrow$ SetModelMask $(W_{t-1},p_2)$
                \STATE $W^c_{t} \leftarrow$ Quantization$(W_{t}^c,\omega)$
                \STATE $W^c_{t} \leftarrow$ ClientLocalTrain$(W^c_{t},c)$
            \ENDFOR
            \STATE $W_{t} \leftarrow \sum_{c \in S} \frac{|\mathcal{D}_k|}{\sum_{k=1}^{N}|\mathcal{D}_k|} W_{t}^c$
        \ENDFOR
        \STATE
        \STATE \textbf{ClientLocalTrain($c,W^c$):}
            \STATE $\mathcal{B}\leftarrow$ (split dataset $D_c$ into batches)
            \FOR{each local epoch $e = 0,1,2,...E$}
               \FOR{each batch $b \in \mathcal{B}$}
                    \STATE $W^c\leftarrow W^c - \eta \nabla \mathcal{L}(W^c;b)$
                \ENDFOR
            \ENDFOR
            \STATE \textbf{return} $W^c$ to server
    \end{algorithmic}
\end{algorithm}

\section{Datasets}
\label{sec:appendix-datasets}
\subsection{Datasets details}
\begin{table}[htb]
  \centering
  \begin{tabular}{lccc}
    \hline
    \textbf{Datasets} & \textbf{\#Train} &\textbf{\#Validation} &\textbf{\#Test} \\
    \hline
    E2E   & 42,061& 4,672& 4,693 \\
    DART   & 62,659 & 2,768 & 5,097         \\
    DialogSum   &  12,460 & 500 & 1,500          \\
    Viggo    &5,103& 714& 1,083          \\
    \hline
  \end{tabular}
  
  \caption{Statistics for different datasets.}
  \label{tab:datasets}
\end{table}
The E2E dataset comprises a collection of table-to-text generation data, primarily designed for training end-to-end natural language generation systems within the restaurant domain. In contrast, the ViGGO dataset, while also a table-to-text generation resource, is intended to support a broader range of conversational dialogue act types. DART, on the other hand, is an open-domain table-to-text generation dataset, and DialogSum is specifically tailored for the task of dialogue summarization. The specific amount of data contained in each dataset is provided in Table \ref{tab:datasets}.
\section{Baselines}

\begin{itemize}

    \item \textbf{FedAVG}: The original global model is distributed to users, who then train it using their local private data. In each training round, the server aggregates the models trained by different users, without applying any random masks or model quantization during the training. This serves as an upper-bound baseline.
    
    \item \textbf{FedSP}: This is the first method proposed to simultaneously protect model privacy in the federated learning framework for large language models. The FedSP approach relies on distilling a proxy model from the global model prior to initiating the federated learning process. However, this approach traditionally requires access to clients' local datasets during the distillation phase, thereby undermining the fundamental privacy preservation principles inherent in federated learning architectures.

    \item \textbf{FedLPP}: It is a federated learning algorithm improved upon FedAVG to protect model privacy. The main improvement lies in the transmission of quantized LoRA between the server and users. However, there is still a possibility that users may have access to the server-side model backbone, meaning the privacy of the global model is not fully protected. This method serves as the primary comparison point in our study.

\end{itemize}

\begin{table*}[htb]
\centering
\begin{tabular}{llccccc}
\toprule[1.3pt]
\multicolumn{2}{c}{\textbf{Method}} & \textbf{BLEU} & \textbf{NIST} & \textbf{METEOR} & \textbf{ROUGE-L} & \textbf{CIDEr} \\
\midrule[1.3pt]
\multirow{2}{*}{$p_1 = 0.1,p2 = 0.1$, w/o quantization} & global & 56.71&7.81&43.37&64.84&1.73
 \\
\cline{2 - 7} 
& proxy & 55.99&7.80&42.16&64.08&1.73 \\
\hline
\multirow{2}{*}{$p_1 = 0.2, p_2=0.2$, w/o quantization} & global & 55.39 & 7.80 & 39.74 & 62.08 & 1.63 \\
\cline{2 - 7} 
& proxy & 53.72 & 7.56 & 39.12 & 61.28 & 1.45 \\
\hline
\multirow{2}{*}{w/o mask, $\omega = 2$} & global & 54.41 & 7.54 & 34.27 & 58.20 & 1.46 \\
\cline{2 - 7} 
& proxy & 32.78 & 0.90 & 23.44 & 51.67 & 0.72 \\
\hline
\multirow{2}{*}{w/o mask, $\omega = 4$} & global & 56.94&7.80&35.43&59.65&1.66 \\
\cline{2 - 7} 
& proxy & 48.89 & 3.63 & 27.13 & 56.11 & 1.07 \\
\hline
\multirow{2}{*}{$p_1= 0.1,p_2=0.1, \omega = 2$} & global & 56.88&7.87&39.55&61.93&1.77 \\
\cline{2 - 7} 
& proxy & 40.02&2.34&25.05&53.43&0.84 \\
\hline
\multirow{2}{*}{$p_1= 0.1,p_2=0.1, \omega = 4$} &global & 55.46&7.74&34.91&58.48&1.54 \\
\cline{2 - 7} 
& proxy & 46.41&3.04&26.52&55.36&1.01\\
\hline
\multirow{2}{*}{$p_1 = 0.2, p_2=0.2, \omega = 4$} & global & 52.78&7.24&32.54&57.23&1.45 \\
\cline{2 - 7} 
& proxy & 40.64&3.07&26.65&53.55&0.97 \\
\bottomrule[1.3pt]
\end{tabular}
\caption{This table presents supplementary comparison experiments with varying mask and quantization levels. The server and client mask ratios are denoted as $p_1$ and $p_2$, respectively, with detailed configurations provided in Section \ref{sec:random-mask}. "w/o mask" indicates the absence of random masking, while "w/o quantization" refers to the absence of quantization. $\omega$ denotes the quantization bit width.}
\label{tab:w/o_mask}
\end{table*}

\section{Experiments}
\label{sec:appendix-experiments}
\subsection{Implementation Details}
To ensure fair and consistent comparisons, we conducted hyperparameter searches for each dataset and method. The best model was selected based on the validation loss, and the corresponding test set results were reported. We primarily evaluated the performance of FedQSN across three quantization levels, specifically with bit width $\omega \in \{1,2,3,4\}$, while maintaining a fixed block size of 256. For the random masking level 
$p$, we selected values from the set $\{0.05, 0.1, 0.15, 0.2\}$.

For FedSP, the prefix length was chosen from $\{40, 80, 160\}$, and the number of layers in the proxy model was selected from \{1, 4, 8\}. For all methods, the learning rate was chosen from $\{10^{-4}, 3\times10^{-4}, 10^{-3}\}$, with a batch size fixed at 16. The number of local training epochs per client in each round was selected from $\{1, 3, 5 \}$, with a total of 10 communication rounds. We implemented the proposed approach, FedQSN, along with the baseline methods, using Hugging Face Transformers~\cite{wolf2020transformers}. All experiments were conducted on a single server equipped with four NVIDIA GeForce RTX 3090 GPUs, each with 24 GB of memory.
\subsection{Parameter Similarity Analysis}
\label{sec:para-similarity}
To evaluate the effectiveness of our method in protecting model privacy, we assess not only the performance gap between the global model and the proxy model, but also the cosine similarity between their parameters. A lower degree of parameter similarity suggests enhanced protection of the model’s proprietary knowledge or intellectual property. As shown in Table \ref{tab:para_similarity}, we compare the parameter similarity between the global and proxy models for FedLPP and FedQSN, using a fixed quantization bit-width of 2. The experiments are conducted with the GPT2-medium model.

The experimental results demonstrate that FedQSN achieves significantly lower parameter similarity compared to FedLPP. This can be attributed to the design of FedLPP, in which the full backbone model is distributed to clients prior to training, and during training, only the quantized LoRA matrices are exchanged between the server and clients. As a result, FedLPP offers relatively weaker protection of the global model’s privacy.

\subsection{Additional ablation study}
In Table \ref{tab:w/o_mask}, we present comparative experiments that examine the effectiveness of applying either masking or quantization alone. The results demonstrate that applying quantization alone leads to a greater performance loss for the global model, but provides stronger protection for the model, as evidenced by a larger gap between the global model and the proxy model. On the other hand, applying only masking results in less performance degradation for the global model, but offers slightly weaker model protection. When one of the methods (quantization or masking) is kept fixed and the other is adjusted, both approaches enhance model privacy protection.

In conclusion, our experiments show that both masking and quantization contribute to improving model privacy, with each offering distinct trade-offs in terms of performance and protection effectiveness.

\end{document}